\documentclass[10pt,twocolumn,letterpaper]{article}

\usepackage{iccv}
\usepackage{times}
\usepackage{epsfig}
\usepackage{graphicx}
\usepackage{amsmath}
\usepackage{amssymb}
\usepackage{multirow}
\usepackage{stfloats}
\usepackage{booktabs}
\usepackage{tabularx}
\usepackage{paralist,lipsum}

\usepackage[pagebackref=true,breaklinks=true,letterpaper=true,colorlinks,bookmarks=false]{hyperref}

\iccvfinalcopy 

\ificcvfinal\fi

\begin{document}

\title{EqGAN: Feature Equalization Fusion for Few-shot Image Generation}

\author{Yingbo Zhou, Zhihao Yue, Yutong Ye, Pengyu Zhang, Xian Wei, Mingsong Chen 
}

\maketitle
\ificcvfinal\thispagestyle{empty}\fi

\begin{abstract} 
   Due to the absence of fine structure and texture information, existing fusion-based few-shot image generation methods suffer from unsatisfactory generation quality and diversity. 
   To address this problem, we propose a novel feature \textbf{Eq}ualization fusion \textbf{G}enerative \textbf{A}dversarial \textbf{N}etwork (\textbf{EqGAN}) for few-shot image generation. 
   Unlike existing fusion strategies that rely on either deep features or local representations, we design two separate branches to fuse structures and textures by disentangling encoded features into shallow and deep contents. 
   To refine image contents at all feature levels, we equalize the fused structure and texture semantics at different scales and supplement the decoder with richer information by skip connections. 
   Moreover, since the fused structures and textures may be inconsistent with each other, we devise a consistent equalization loss between the equalized features and the intermediate output of the decoder to further align the semantics.
   Comprehensive experiments on three public datasets demonstrate that, EqGAN not only significantly improves the FID scores (by up to 32.7\%) and LPIPS scores (by up to 4.19\%) of generated images, but also outperforms the state-of-the-arts in terms of accuracy (by up to 1.97\%) for downstream classification tasks.
\end{abstract}

\section{Introduction}
Along with the prosperity of deep learning techniques, Few-Shot Learning (FSL)~\cite{WangYKN20, abs-2205-06743} has shown outstanding performance in a variety of visual tasks with insufficient data (e.g., few-shot image recognition~\cite{PengLZLQT19, JiangZLS0M23}, few-shot image segmentation~\cite{WangLZZF19, TangLSYX21}, and few-shot image generation~\cite{ZhaoDHC22, Gu0H0021, YangWCF22}). 
However, due to the difficulty in obtaining the data distribution from limited images, FSL needs to learn how to learn knowledge effectively by using meta-learning~\cite{HospedalesAMS22}.
To increase the amount of data and accommodate visual applications in limited-data scenarios, few-shot image generation is regarded as a data-augmentation approach to generate more new images, which can bring benefits to downstream tasks such as low-data detection~\cite{KangLWYFD19, WuL0W20} and few-shot classification~\cite{VinyalsBLKW16, SnellSZ17}.

\begin{figure}[htbp]
\begin{center}
   \includegraphics[width=0.85\columnwidth]{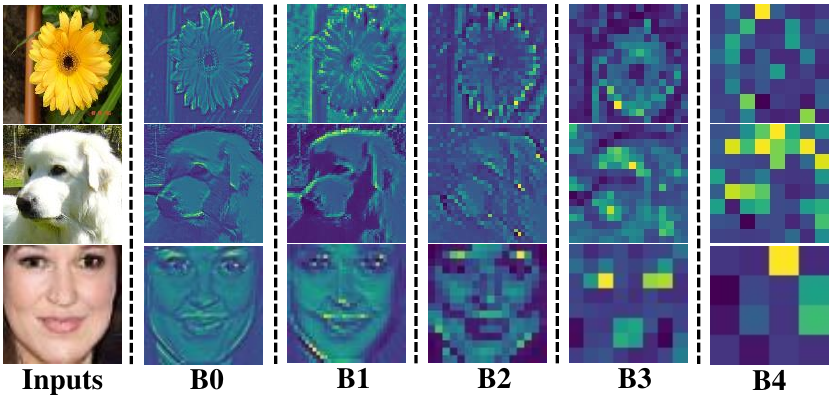}
\end{center}
\vspace{-0.2cm}
   \caption{\textbf{Visualization of feature maps in the encoder.} The inputs are real images from  public datasets. The notations of ``B0'', ``B1'', ``B2'', ``B3'' and ``B4'' indicate the feature maps corresponding to the five shallow-to-deep blocks in the encoder, respectively.}
\label{fig:vi}
\vspace{-0.65cm}
\end{figure}

Aiming to produce new images for a target category, few-shot image generation is expected to learn transferable generation ability from seen categories with sufficient labeled training images. 
Concretely, a generative model is firstly trained on an auxiliary dataset (i.e., seen categories) by the episodic training mechanism~\cite{VinyalsBLKW16}. 
Then, the learned generative model is used to yield diverse and realistic images for a target category (i.e., unseen category) with a few images.  
So far, existing few-shot image generation methods can be mainly classified into three categories: 
i) optimization-based methods~\cite{abs-1901-02199, abs-2001-00576} that 
focus on learning an initialized base model with meta-learning paradigms for unconditional image generation tasks, resulting in low-quality generated images;
ii) transformation-based methods~\cite{abs-1711-04340, HongNZZ22} that introduce intra-category transformations as extra knowledge to unseen samples to generate more images for the unseen category, requiring complicated image transformations with unstable training processes; and iii) fusion-based methods~\cite{BartunovV18, Hong00ZF020, HongN0020a, Gu0H0021, YangWCF22} that fuse several given images of the same category with different fusion strategies to generate new images, obtaining limited improvements in both generation quality and diversity. 
In this paper, we concentrate on exploring a more rational fusion strategy for better image generation. 

The essence of fusion-based few-shot generation is to build a class-invariant mapping from a handful of images with the same label. 
However, previous fusion strategies cannot efficiently learn a generalizable mapping with semantic alignment.
To explain the inadequacy of existing fusion strategies, Figure~\ref{fig:vi} presents an example that  visualizes the encoded feature maps of different convolution blocks.
From this figure, we can observe that the feature maps from B0, B1, and B2 mainly focus on texture information while the feature maps from B3 and B4 pay close attention to structure information, indicating
that the feature maps from different blocks of the encoder contain different semantic information (i.e., structure semantics from the deep blocks and texture details from the shallow blocks). 
Since existing fusion strategies directly fuse the encoded features at the global content~\cite{BartunovV18, Hong00ZF020, HongN0020a} or local representation~\cite{Gu0H0021, YangWCF22} level, they usually lead to semantic entanglement in the fused features.
Therefore, \emph{how to implement a more fine-grained fusion strategy to avoid semantic entanglement} is becoming one of the major challenges for fusion-based methods.
To address  this problem, this paper makes three major contributions as follows:
\begin{itemize}
    \item We propose a novel few-shot image generation framework called \textbf{EqGAN} (feature \textbf{Eq}ualization fusion \textbf{G}enerative \textbf{A}dversarial \textbf{N}etwork), which can fuse features in the structure and texture branches to achieve better generation quality and diversity.
    \item We design a feature equalization fusion strategy to train the model with more refined guidance, which can equalize the fused features at multi-scale levels to mitigate semantic entanglement.
    \item We devise a consistent equalization loss to strengthen the interaction between the encoder and the decoder, which further guarantees the consistency of generated images in semantic features. 
\end{itemize}

Comprehensive experiments on three well-known datasets show that our proposed EqGAN can not only outperform the state-of-the-art approaches for few-shot image generation, but also achieve significant improvements in the accuracy of downstream classification tasks.

\section{Related Work}


\textbf{Generative Adversarial Networks.} 
Generative Adversarial Networks (GANs)~\cite{BrockDS19, KarrasLA19, LiuZSE2021} have demonstrated powerful performance in various visual domains since the pioneering work in~\cite{Goodfellow14}, including image generation~\cite{LiuZSE2021, ShimHBH22, abs-2106-14490}, image harmonization~\cite{Cong0NLLL020, GuoGZGZD21, JiangZ0WLSCAKW21}, image inpainting~\cite{LiuJSHY20, GuoY021, SuinP021}, and image-to-image translation~\cite{RichardsonAPNAS21, BaekCUYS21, WangM0LR21}. 
Although GANs become
prevalent in these fields owing to their excellent ability to fit a data distribution through adversarial learning, training a GAN for great performance often requires an unlimited supply of training images to obtain a good data distribution.
Worse still, the discriminator tends to overfit in  case of scarce data, which makes the model susceptible to divergence and leads to poor results. 
Recently, the mainstream GAN-based methods utilize advanced data augmentation techniques ~\cite{JiangDWL21, ZhaoLLZ020} or regularization techniques~\cite{TsengJL0Y21, WuSTC21} to guide the discriminator to avoid overfitting in  scenarios with limited data. Different from existing mainstream methods, this paper presents a few-shot learning paradigm to alleviate the overfitting problem of GAN models. Given a few images from an unseen category, our approach tries to train  a GAN model to generate realistic and diverse images for the target category. 

\textbf{Few-shot Image Generation.}
To imitate the ability of learning to learn from a few observations like human beings, few-shot image generation tasks attempt to produce new images for an unseen category with limited data. 
So far, existing few-shot image generation methods can be  classified
into three categories: i) optimization-based methods~\cite{abs-1901-02199, abs-2001-00576}, ii) transformation-based methods~\cite{abs-1711-04340, HongNZZ22}, and iii) fusion-based methods~\cite{BartunovV18, Hong00ZF020, HongN0020a, Gu0H0021, YangWCF22}. 
The first category combines GANs with different meta-learning optimization algorithms. 
For example, FIGR~\cite{abs-1901-02199} and DAWSON~\cite{abs-2001-00576} use Reptile~\cite{abs-1803-02999} and MAML~\cite{FinnAL17} to generate new images, respectively. However, both methods have poor authenticity in image generation. 
For the second category, DAGAN~\cite{abs-1711-04340} implements image transformation by combining the projected latent codes while DeltaGAN~\cite{HongNZZ22} captures the intra-category transformation by designing the reconstruction subnetwork. However, the end-to-end training of image transformations is very unstable and the generated images continually appear low-quality. 
The third category usually fuses the features by matching the global contents or interpolating the local representations to generate images with several conditional samples.
For instance, GMN~\cite{BartunovV18} and MatchingGAN~\cite{HongN0020a} fulfill the rough fusion by combining the matching network with VAE~\cite{Doersch16} and GAN, respectively, while F2GAN~\cite{Hong00ZF020} adds a non-local attention fusion module with a fuse-and-fill strategy to improve MatchingGAN. 
However, these  fusion strategies are prone to  imprecise generation quality and limited diversity. 
More recently, LoFGAN~\cite{Gu0H0021} introduces a local fusion strategy with local representations to generate images with fewer aliasing artifacts, and WaveGAN~\cite{YangWCF22} employs high-frequency signals with skip connections in the generator to further promote the performance of LoFGAN.

Although LoFGAN utilizes a fine-grained fusion strategy to generate new images for few-shot image generation tasks, the details of generated images still blur due to the entanglement of fused features. 
Similarly, WaveGAN only disentangles the encoded features into multiple frequency components for the base image, which barely makes a difference in generation diversity. 
To the best of our knowledge, this paper is the first attempt 
that considers
semantic entanglement in few-shot image generation.
By disentangling the encoded features and using the feature equalization operation to fuse the structure semantics and texture details, our proposed method can not only improve generation quality and diversity, but also have better accuracy for downstream classification tasks. 

\begin{figure*}
\begin{center}
\includegraphics[width=1.95\columnwidth]{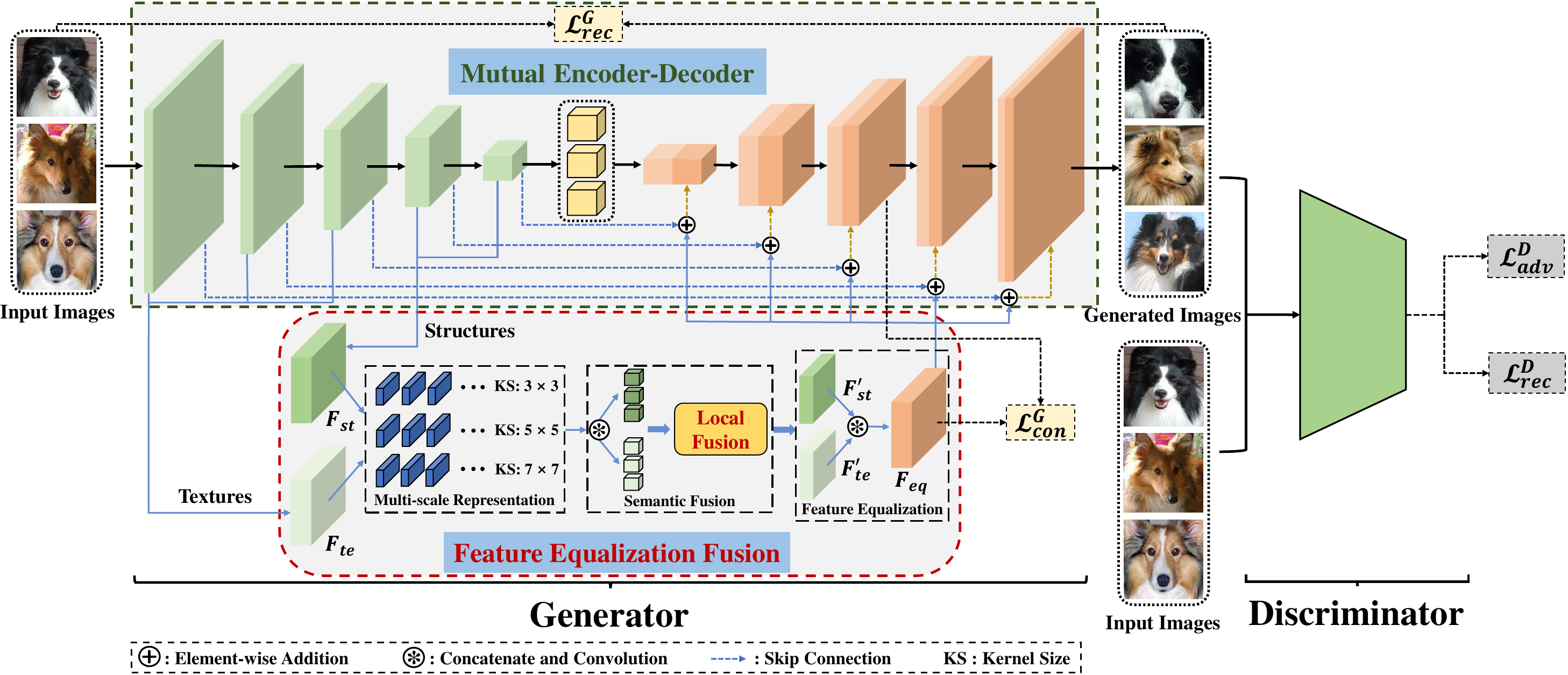}
\end{center}
\vspace{-0.2cm}
   \caption{\textbf{Overall framework of our EqGAN.} Our EqGAN consists of an encoder-decoder generator, a discriminator, and a feature equalization fusion module, where the feature equalization fusion module fuses encoded features in structure and texture branches to provide richer and more consistent information for improving synthesis quality. Local fusion is a basic block for our fusion strategy.}
\label{fig:framework}
\vspace{-0.3cm}
\end{figure*}

\section{Methodology}

\subsection{Overview}

\textbf{Problem Definition.} Given $K$ samples from an unseen category, our intention is to generate new realistic and diverse images for the same category. The number of samples $K$ is generally small (i.e., 3 or 5) and defines a $K$-shot image generation task. To accomplish this task, we need to divide a complete dataset into two parts, i.e.,  seen categories $\mathbb{C}_s$ and unseen categories $\mathbb{C}_u$, where $\mathbb{C}_s$ and $\mathbb{C}_u$ have no intersection in label space. In the training phase, we train the model with hundreds of $K$-shot image generation tasks sampled from $\mathbb{C}_s$, expecting our model to learn a transferable mapping and generate new images for a seen category. In the testing phase, a handful of images from one category in $\mathbb{C}_u$ are fed into the trained model to synthesize new images for this unseen category.

\textbf{Overall Framework.} As shown in Figure~\ref{fig:framework}, our EqGAN consists of a generator with a feature equalization fusion module and a discriminator. The generator $G$ adopts a mutual encoder-decoder architecture with skip connections. The encoder and the decoder are symmetric, both are made up of five convolutional blocks. The encoder first extracts deep features from the input images and directly transfers the encoded features to the decoder. Meanwhile, we reorganize the encoded features from the shallow layer and deep layer of the encoder as texture and structure branches, respectively. Then, both branches are equalized by our feature equalization fusion module. Finally, the equalized features at different feature levels supplement the decoder via skip connections to generate images. For the discriminator $D$, we feed input images and generated images into it for adversarial training. Next, we will elaborate on our encoder-decoder architecture, the feature equalization fusion module and the design of optimization objectives.

\subsection{Mutual Encoder-Decoder}
Considering that existing fusion-based architectures cannot effectively fuse the features of different images, we design a mutual encoder-decoder generator to get satisfactory images.
Inspired by  U-Net~\cite{RonnebergerFB15}, we build skip connections between the encoder and the decoder to retain shallow and deep semantic features. 
To make the generator $G$ effectively learn more fused semantic information, we perform element-wise addition operations before connecting the encoder with the decoder.
Similar to ResNet~\cite{HeZRS16}, we utilize element-wise addition operations to bring more content in each feature dimension for the decoder.  

For ease of description, we denote the encoder blocks (resp., decoder) from shallow to deep as $E_1$ (resp., $H_5$) to $E_5$ (resp., $H_1$). For $r=\{1, 2, 3, 4, 5\}$, the $r^{th}$ skip connection directs the output from the $r^{th}$ block of the encoder to the output from the $r^{th}$ block of the decoder. 
In Figure~\ref{fig:framework}, we omit the local fusion process in our encoder-decoder for the convenience of observation.
Given a small set of  input images $X=\{x_1, ..., x_K \}$ and the concatenation operation $[\cdot]$, we  formulate the relationship between the encoder $E$ and the decoder $H$ as:
\begin{equation}
    {H}_{i+1}(X) = {\left [{H}_{i}(X), {E}_{i}(X_{b}) \oplus EqBlock_{i}(X)\right ]},
\end{equation}
where $X_b$ denotes
a base image randomly selected from $X$,  $i$ is an integer ranging from $1$ to $5$, ${H}_{i}(\cdot)$ and ${E}_{i}(\cdot)$ indicate the output in the $i^{th}$ block of the encoder and the decoder, respectively. Here, $\oplus$ represents element-wise addition operations, and $EqBlock_{i}(X)$ denotes the feature equalization fusion process, which will be detailed later. 

\subsection{Feature Equalization Fusion}

Since different layers of the model focus on different semantics, we can obtain multi-layer structure features $F_{st}$ and texture features $F_{te}$ as shown in Figure~\ref{fig:framework}. 
To be concrete, we regard the encoded features obtained by the first three convolutional blocks of the encoder as textures, and the encoded features from the last two convolutional blocks as structures.
Considering that the encoded feature maps from different convolutional blocks are different, we transform them to the same size using multi-scale convolution operations and then concatenate them accordingly. 
The specific functions for $F_{st}$ and $F_{te}$ are defined as follows:
\begin{equation}
    \begin{aligned}
        F_{te} & = \sum_{i}ConvDown_{i}(E_{i}(X)),~~~{(i=1,2,3)}, \\
        F_{st} & = \sum_{i}ConvUp_{i}(E_{i}(X)),~~~~~~~~{(i=4,5)}.
    \end{aligned}
\end{equation}
where $ConvDown(\cdot)$ and $ConvUp(\cdot)$ denote the down-convolution and up-convolution operations, respectively. The detailed fusion process involves three major steps, i.e.,  multi-scale representation, semantic fusion, and feature equalization. 


\begin{figure}[t]
\begin{center}
   \includegraphics[width=1.0\columnwidth]{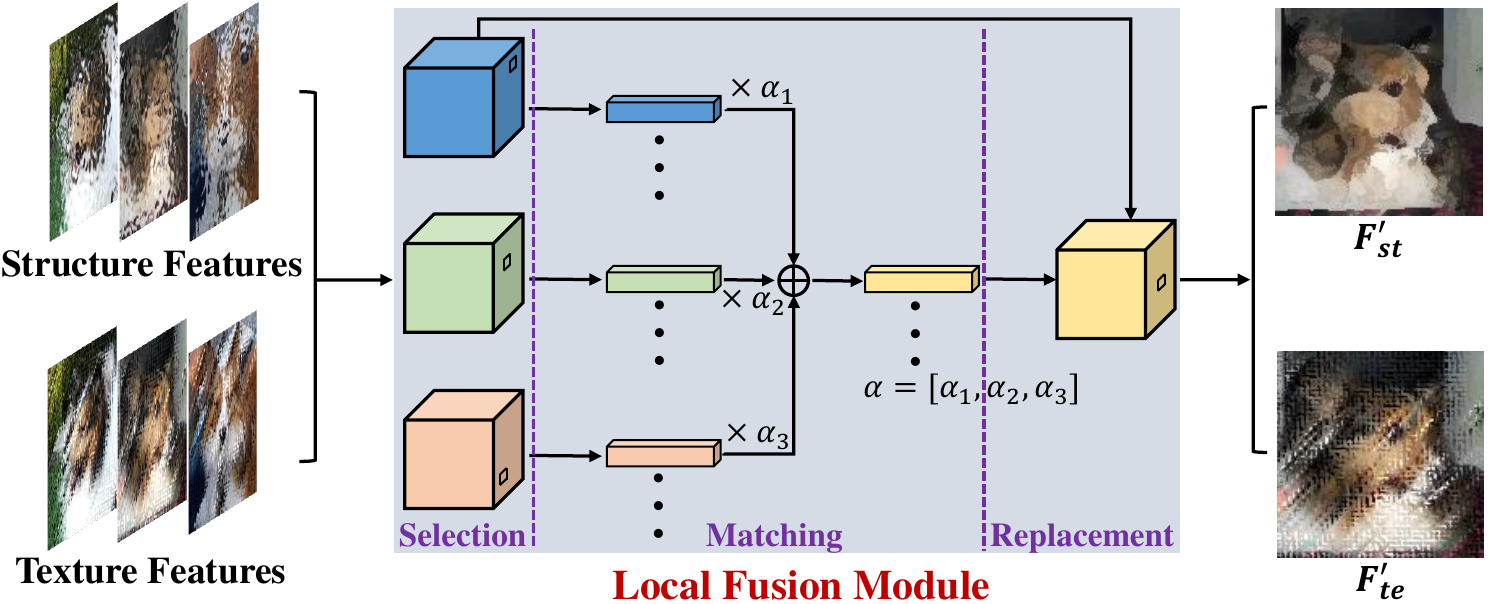}
\end{center}
\vspace{-0.2cm}
   \caption{\textbf{Implementation of our semantic fusion process.} 
   The blue cube indicates  base features, the green and orange cubes denote  reference features, and the yellow cube is
   fused features. 
   Here we illustrate the fusion process with $3$-shot image generation.}
\label{fig:lfm}
\vspace{-0.3cm}
\end{figure}


\textbf{Multi-scale Representation.}  
After the reorganization of encoded features, we use two branches (i.e., the structure branch and the texture branch) to obtain the  multi-scale representations of corresponding structures and textures. 
In each branch, there are three parallel streams to capture multi-grained feature representations, which we depict in Figure~\ref{fig:framework} with one multi-scale convolution module. 
Each stream consists of 5 convolutions with the same kernel size while the kernel size differs among different streams (i.e. $3\times 3$, $5\times 5$, $7\times 7$). 
By using different kernel sizes, we can adaptively perceive more detailed information.

\textbf{Semantic Fusion.} In the semantic fusion stage, our strategy is based on the local fusion module of LoFGAN~\cite{Gu0H0021}, which can be seen in Figure~\ref{fig:lfm}. 
Here we give a brief introduction to the local fusion module. 
Given the input images $X = \{ {x_1, ..., x_K} \}$, the encoder extract deep features $F = E(X) \in \mathbb{R}^{K\times w \times h \times c}$. 
Then, the model randomly selects one feature from $F$ as a base feature $f_{base} \in \mathbb{R}^{w \times h \times c}$ and views the rest of $K-1$  features as reference features $F_{ref} \in \mathbb{R}^{(K-1)\times w \times h \times c}$. 
For each reference feature $f^{i}_{ref} \in F_{ref}$, we calculate the  similarity map $\mathcal{M}$ as follows:
\begin{equation}
    \mathcal{M}^{(i,j)} = g(f^{(i)}_{base}, f^{(j)}_{ref}),
\end{equation}
where $i \in \{1, ..., n\}$, $j \in \{1, ..., h\times w\}$, and $g$ is a similarity metric. 
According to the similarity map, we can find the most similar semantic information for each position in $f_{base}$. 
In this way, LoFGAN can match the semantic similarity map to fuse the local representations and replaces the closest base feature with the fused features.

Unlike LoFGAN that uses  encoded deep features to complete local representation fusion, we treat the structures and textures as two separate units for fusing. 
In other words, we replace the local representations with more fine-grained semantics (i.e., the structures and textures) during the fusion process.
To merge the multi-scale texture feature maps and structure feature maps, we concatenate and map them to the same size by $1\times 1$ convolutions. 
We denote the input of the structure branch and the texture branch as structure features and texture features in Figure~\ref{fig:lfm}, respectively. 
The fused structure features $F^{'}_{st}$ and texture features $F^{'}_{te}$ in semantic fusion process are defined as:
\begin{equation}
    \begin{aligned}
        F^{'}_{te} & = \sum_{i=1, ..., K}\alpha_i \cdot MultiConv(F_{te}(x_i)), \\
        F^{'}_{st} & = \sum_{i=1, ..., K}\alpha_i \cdot MultiConv(F_{st}(x_i)),
    \end{aligned}
\end{equation}
where $\sum_{i=1}^{K}\alpha_i=1$, $\alpha_i\geq 0$, and $MultiConv(\cdot)$ denotes the multi-scale representation operations.

\textbf{Feature Equalization.} The fused semantics in $F^{'}_{st}$ and $F^{'}_{te}$ may be inconsistent to reflect the generation of fine-grained structures and textures. To ensure that $F^{'}_{st}$ and $F^{'}_{te}$ are compatible with each other, we perform concatenation operations on them first and make a simple fusion to obtain the equalized features $F_{eq}$ via a $1 \times 1$ convolutional layer. Meanwhile, we formulate a consistent equalization loss between the output in the middle layer of the decoder and the equalized features $F_{eq}$, which can further mitigate blur and artifacts within and around the fused regions. The detailed form of our consistent equalization loss will be described in the following section.

\subsection{Optimization Objective Design}

In the whole training process, we need to optimize our model in two parts: the generator $G$ and the discriminator $D$. Let $X = \{ x_1, ..., x_K \} $ and $\hat{X} = \{ \hat{x}_1, ..., \hat{x}_K \}$ denote the sets of input images and  generated images, respectively. The labels of seen categories in $X$ are represented by $C(X)$. The generator $G$ and the discriminator $D$ are optimized alternately in an adversarial manner by the following losses.

\textbf{Consistent Equalization Loss.} To strengthen the interaction between the encoder and the decoder, we devise a consistent equalization loss on the intermediate output of the decoder $F_{H3}$ and the equalized features $F_{eq}$, which can relieve the inconsistency of the fused information. Let ${\Vert \cdot \Vert}_1$ denote the $L_1$ norm, and the consistent equalization loss is defined as follows:
\begin{equation}
    \mathcal{L}^{G}_{con} = {\Vert F_{eq} - F_{H3} \Vert}_1.
\end{equation}

\textbf{Local Reconstruction Loss.} To reproduce the feature-level local fusion procedure, we adopt the local reconstruction loss~\cite{Gu0H0021} to constrain the generator $G$ as below:
\begin{equation}
    \mathcal{L}^{G}_{rec} = {\Vert \hat{X} - LFM(X, \alpha) \Vert}_1,
\end{equation}
where $\alpha$ is a random coefficient vector to control the weight of the fused features. 

\textbf{Adversarial Loss.} According to~\cite{Gu0H0021}, we train the generator $G$ and the discriminator $D$ alternately based on the hinge version loss~\cite{OdenaOS17}, which can be formulated as: 
\begin{equation}
    \begin{aligned}
        \mathcal{L}^{D}_{adv} & = max\{ 0, 1 - D(x_i) \} + max\{ 0, 1 + D(\hat{x}_{i}) \}, \\
        \mathcal{L}^{G}_{adv} & = - D(\hat{x}_{i}), ~~~~~~~i\in \{1, ..., K\}.
    \end{aligned}
\end{equation}

\textbf{Classification Loss.} To constrain the model to produce images with one specific label, we add an auxiliary classifier to the model according to ACGAN~\cite{OdenaOS17}. 
The classification loss is defined as:
\begin{equation}
    \begin{aligned}
        \mathcal{L}^{D}_{cls} & = - log{P(c(X) \vert X)}, \\
        \mathcal{L}^{G}_{cls} & = - log{P(c(X) \vert {\hat{X}})}.
    \end{aligned}
\end{equation}

\textbf{Total Losses.} The overall objective function of the proposed network is optimized as follows:
\begin{equation}
    \begin{aligned}
        \mathcal{L}_{G} & = \mathcal{L}^{G}_{adv} + {\lambda\ ^{G}_{cls}}\mathcal{L}^{G}_{cls} + {\lambda\ ^{G}_{rec}}\mathcal{L}^{G}_{rec} + {\lambda\ ^{G}_{con}}\mathcal{L}^{G}_{con}, \\
        \mathcal{L}_{D} & = \mathcal{L}^{D}_{adv} + {\lambda\ ^{D}_{cls}}\mathcal{L}^{D}_{cls},
    \end{aligned}
\end{equation}
where $\lambda\ ^{G}_{cls}$, $\lambda\ ^{G}_{rec}$, $\lambda\ ^{G}_{con}$, and $\lambda\ ^{D}_{cls}$ are the trade-off hyper-parameters to balance the weights of the total losses. 

\section{Experiments}
To evaluate the effectiveness of our method, we implemented EqGAN based on LoFGAN~\cite{Gu0H0021}. 
All the experimental results were obtained using PyTorch 1.13.0 on an Ubuntu operating system with one 3.7GHz Intel CPU, 32GB memory, and NVIDIA RTX 3090TI GPU.
We designed comprehensive experiments to answer the following three research questions:

\textbf{RQ1 (Superiority of generated images):} 
How about the quality and diversity of images generated by EqGAN compared with the ones generated
by the state-of-the-arts?

\textbf{RQ2 (Effectiveness of fusion
components):} 
What is the impact of each component of our proposed fusion strategy on the images generated by EqGAN?

\textbf{RQ3 (Benefit to  downstream classification):} 
How much can  downstream classification tasks benefit from the images generated by EqGAN?

\subsection{Experimental Settings}
\textbf{Hyperparameters.}
In the training process, we trained our model for $100,000$ iterations with Adam optimizer~\cite{KingmaB14}. In the first $50,000$ iterations, the fixed learning rate of the network is set to $0.0001$. In the rest $50,000$ iterations, the learning rate is linearly decayed to $0$. In each iteration, the batch size is $8$, and we need to randomly sample eight $K$-shot image generation tasks from $\mathbb{C}_s$. We set $\lambda\ ^{G}_{cls} = \lambda\ ^{D}_{cls} = \lambda\ ^{G}_{con} = 1$, and $\lambda\ ^{G}_{rec} = 0.5$. In the testing process, we used the final checkpoint to produce new images with $\mathbb{C}_u$. 

\textbf{Datasets.} 
We evaluated our method on three well-known datasets, namely Flower~\cite{NilsbackZ08}, Animal Faces~\cite{0001HMKALK19} and VGGFace~\cite{CaoSXPZ18}.
According to \cite{Gu0H0021} and \cite{YangWCF22}, we set the split of datasets as shown in Table~\ref{table:data}.
\begin{table}[!h]
\begin{center}
\setlength{\tabcolsep}{2.0mm}{
\tabcolsep=0.15cm
\begin{tabular}{lcccc}
\toprule[1pt]
Datasets     & $\mathbb{C}_{t}$ (\#) & $\mathbb{C}_{s}$ (\#) & $\mathbb{C}_u$ (\#) & N/C (\#) \\ \hline
Flower~\cite{NilsbackZ08}       & 102                & 85                & 17                  & 40                \\
Animal Faces~\cite{0001HMKALK19} & 149                & 119               & 30                  & 100               \\
VGGFace~\cite{CaoSXPZ18}      & 2354               & 1802              & 552                 & 100               \\ \bottomrule[1pt]
\end{tabular}}
\end{center}
\vspace{-0.1cm}
\caption{\textbf{The split of datasets.} $\mathbb{C}_{t}$ denotes the number of categories in each dataset, $\mathbb{C}_{s}$  and $\mathbb{C}_u$ represent the number of seen categories and unseen categories in each dataset, respectively. N/C indicates the number of images selected from each category.}
\label{table:data}
\vspace{-0.2cm}
\end{table}


\textbf{Baselines.} We compared our EqGAN with several few-shot image generation methods, namely FIGR~\cite{abs-1901-02199}, DAWSON~\cite{abs-2001-00576}, DAGAN~\cite{abs-1711-04340}, GMN~\cite{BartunovV18}, MatchingGAN~\cite{HongN0020a}, F2GAN~\cite{Hong00ZF020}, DeltaGAN~\cite{HongNZZ22}, LoFGAN~\cite{Gu0H0021} and WaveGAN~\cite{YangWCF22}. To make a fair comparison with the state-of-the-art fusion-based methods, we reimplemented LoFGAN and WaveGAN with the same network architecture and hyper-parameters,  denoted as LoFGAN$\dagger$ and WaveGAN$\dagger$, respectively. Note that, all the above methods were evaluated under the same settings.

\textbf{Metrics.} We evaluated the quality of generated images based on  two commonly used metrics, i.e.,  Fr$\acute{e}$chet Inception Distance (FID)~\cite{HeuselRUNH17} and  Learned Perceptual Image Patch Similarity (LPIPS)~\cite{ZhangIESW18}, where
FID is a measure to calculate the distance between two image feature vectors, and  LPIPS, in line with human perception, is used to measure the difference between two images. The lower the score of both metrics, the more similar the two images are. 

\begin{figure*}[t]
\begin{center}
       \includegraphics[width=2.1\columnwidth]{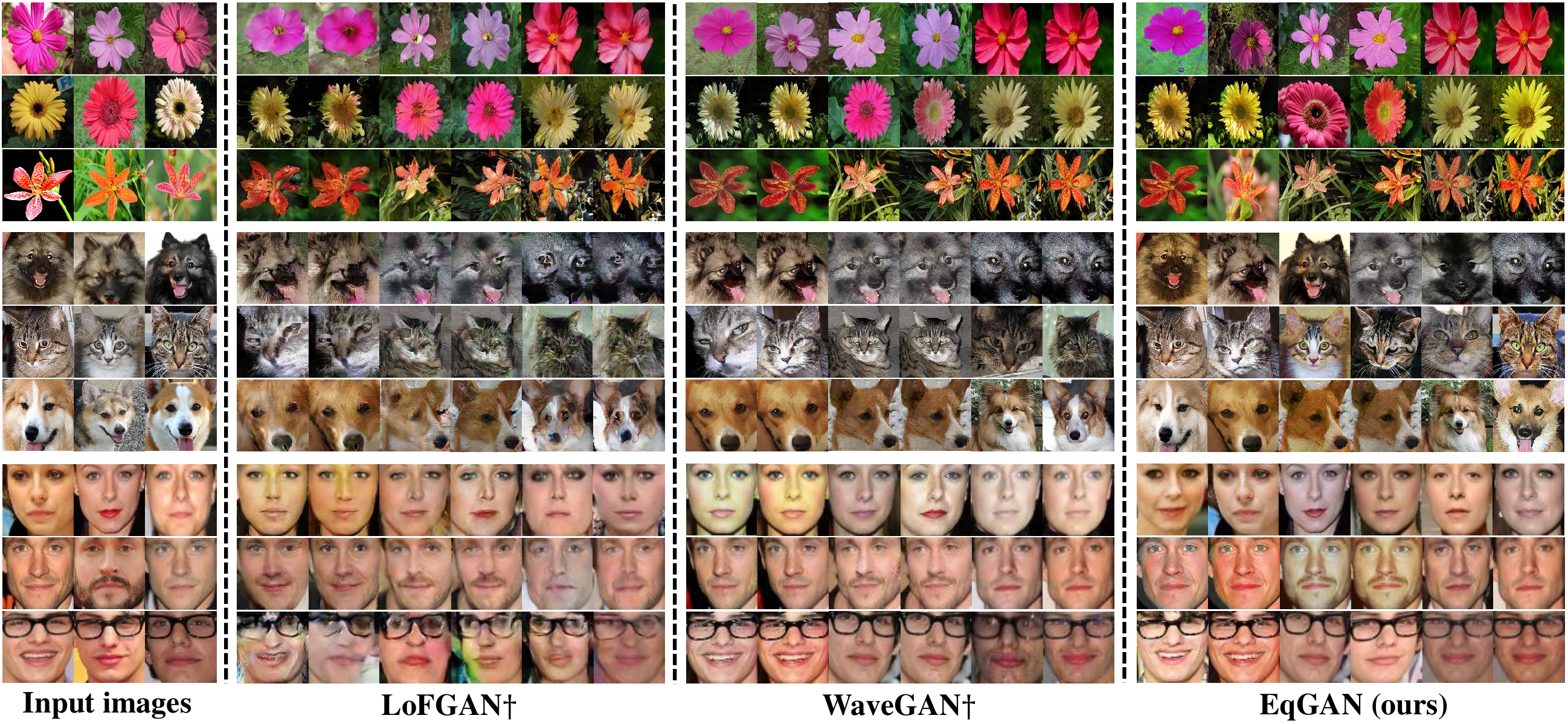}
\end{center}
\vspace{-0.1cm}
   \caption{\textbf{Qualitative comparison results between LoFGAN$\dagger$, WaveGAN$\dagger$ and EqGAN.} The leftmost three columns are input images, and six generated images are selected for each method. The results with symbol $\dagger$ are reimplemented under the same setting as our model.}
\label{fig:visual}
\vspace{0.1cm}
\end{figure*}

\begin{table*}[htbp]
\begin{center}
\setlength{\tabcolsep}{3.5mm}{
\begin{tabular}{lccccccc}
\toprule[1pt]
\multirow{2}{*}{Method} & \multirow{2}{*}{Type} & \multicolumn{2}{c}{Flowers} & \multicolumn{2}{c}{Animal Faces} & \multicolumn{2}{c}{VGGFace} \\
                        &                       & FID $\downarrow$        & LPIPS $\uparrow$        & FID $\downarrow$           & LPIPS $\uparrow$          & FID $\downarrow$        & LPIPS $\uparrow$        \\ \hline
FIGR~\cite{abs-1901-02199}         & Optimization          & 190.12            & 0.0634              & 211.54               & 0.0756                & 139.83            & 0.0834              \\
DAWSON~\cite{abs-2001-00576}         & Optimization          & 188.96            & 0.0583              & 208.68               & 0.0642                & 137.82            & 0.0769              \\
DAGAN~\cite{abs-1711-04340}      & Transformation        & 151.21            & 0.0812              & 155.29               & 0.0892                & 128.34            & 0.0913              \\
GMN~\cite{BartunovV18}      & Fusion        & 200.11            & 0.0743              & 220.45               & 0.0868                & 136.21            & 0.0902              \\
MatchingGAN~\cite{HongN0020a}             & Fusion                & 143.35            & 0.1627              & 148.52               & 0.1514                & 118.62            & 0.1695              \\
F2GAN~\cite{Hong00ZF020}     & Fusion                & 120.48            & 0.2172              & 117.74               & 0.1831                & 109.16            & 0.2125              \\
DeltaGAN~\cite{HongNZZ22}                & Transformation        & 104.62            & 0.4281              & 87.04               & 0.4642                & 78.35             & 0.3487              \\
LoFGAN~\cite{Gu0H0021}                  & Fusion                & 79.33            & 0.3862              & 112.81               & 0.4964                & 20.31            & 0.2869              \\
WaveGAN~\cite{YangWCF22}                 & Fusion                & 42.17            & 0.3868              & 30.35               & 0.5076                & 4.96            & 0.3255              \\ \hline
\textbf{LoFGAN}$\dagger$                  & Fusion                & 82.70            & 0.3824              & 113.73               & 0.5020                & 20.61           & 0.3071              \\
\textbf{WaveGAN}$\dagger$                 & Fusion                &  \underline{46.19}           &  \underline{0.3865}             &  \textbf{34.33}              &  \underline{0.5049}               &  \underline{5.13}           & \underline{0.3233}              \\
\textbf{EqGAN (ours)}                   & Fusion                &  \textbf{45.15}           & \textbf{0.4027}              &  \underline{34.80}              &  \textbf{0.5130}               & \textbf{3.45}            & \textbf{0.3345}              \\ 
\bottomrule[1pt]
\end{tabular}}
\end{center}
\caption{\textbf{Quantitative comparison results between our model and the baselines.} The results of the first six methods are quoted from the work~\cite{Hong00ZF020} and the next three results are quoted from DeltaGAN~\cite{HongNZZ22}, LoFGAN~\cite{Gu0H0021}, and WaveGAN~\cite{YangWCF22}, respectively. The results with symbol $\dagger$ are reimplemented under the same setting as our model. The best and the second-best results are \textbf{highlighted} and \underline{underlined}, respectively. The symbol $\downarrow$ indicates that lower is better while the symbol $\uparrow$ indicates that higher is better.}
\label{table:qua}
\vspace{-0.2cm}
\end{table*}

\subsection{Results on Generation Performance (RQ1)}
To show the superiority of our method in generating high-quality images, we conducted the following experiments to quantitatively and qualitatively evaluate EqGAN considering the impact of the number of shots.

\textbf{Quantitative Results.} To quantitatively compare EqGAN with baselines, we carried out our experiments under a $3$-way generation setting. We first trained the model with seen categories $\mathbb{C}_s$ and then sampled the data from unseen categories $\mathbb{C}_u$ to validate the generation quality. Following the settings of ~\cite{Gu0H0021} and~\cite{YangWCF22}, we split each unseen category into two parts: $\mathbb{C}^{1}_{u}$ and $\mathbb{C}^{2}_{u}$. The images from $\mathbb{C}^{1}_{u}$ were fed to the trained model to produce new $128$ images for each category, where we denoted the generated images as $\mathbb{C}_{g}$. We calculated the FID and LPIPS scores between $\mathbb{C}_{g}$ and $\mathbb{C}^{2}_{u}$ to analyze the generation performance. 

From Table~\ref{table:qua}, we can find that EqGAN outperforms the baseline models on almost all of the three datasets. 
Since the implementation of our proposed method is based on LoFGAN~\cite{Gu0H0021}, we made a clearer comparison with LoFGAN$\dagger$.
Specifically, the FID scores of EqGAN  are \textbf{45.15}, \textbf{34.80} and \textbf{3.45} on Flower, Animal Faces and VGGFace datasets, respectively, while the FID scores of LoFGAN$\dagger$ on these datasets are 82.70, 113.73 and 20.61, respectively.
Meanwhile, our LPIPS scores on Flower, Animal Faces, and VGGFace datasets are \textbf{0.4027}, \textbf{0.5130}, and \textbf{0.3345}, respectively, which are much better than those of LoFGAN$\dagger$ (i.e., 0.3824, 0.5020 and 0.3071). 
As the state-of-the-art fusion-based method, WaveGAN$\dagger$ achieves a lower FID score (34.33) on Animal Faces dataset, but its FID scores on Flower and VGGFace datasets (i.e., 46.19 and 5.13), and the LPIPS scores on the three datasets (i.e., 0.3865, 0.5049 and 0.3233) are inferior to the ones obtained by EqGAN. 
As an example for VGGFace dataset, EqGAN outperforms WaveGAN$\dagger$ (by \textbf{32.75\%}) in FID score and LPIPS score (by \textbf{3.46\%}), which substantiates the effectiveness of our proposed method.   

\textbf{Qualitative Results.} To qualitatively compare EqGAN with baselines, we visualized the generated results of LoFGAN$\dagger$, WaveGAN$\dagger$, and our EqGAN on all three datasets. From Figure~\ref{fig:visual}, we can observe that the images generated by EqGAN are more realistic and diverse than the ones generated by LoFGAN$\dagger$ and WaveGAN$\dagger$. The images generated by LoFGAN$\dagger$ are fuzzy and distorted, where the details of the generation structure and texture have obvious artifacts. For the images generated by WaveGAN$\dagger$, there is little intra-class difference for the same category, indicating that the generation has poor diversity. Comparatively, our method can produce more authentic images with fewer artifacts, richer details, and more texture and color information. Taking the generated animal face images as examples, the results of LoFGAN$\dagger$ in the eyes, the noses, and the hair of animals are distorted and even misplaced, while WaveGAN$\dagger$ repeatedly produces similar samples with limited diversity. By contrast, animal face images generated by EqGAN have more variation in color, texture, and structure, which demonstrates the superiority of our method in generation quality and diversity.

\begin{figure}[htbp]
\begin{center}
   \includegraphics[width=1.0\columnwidth]{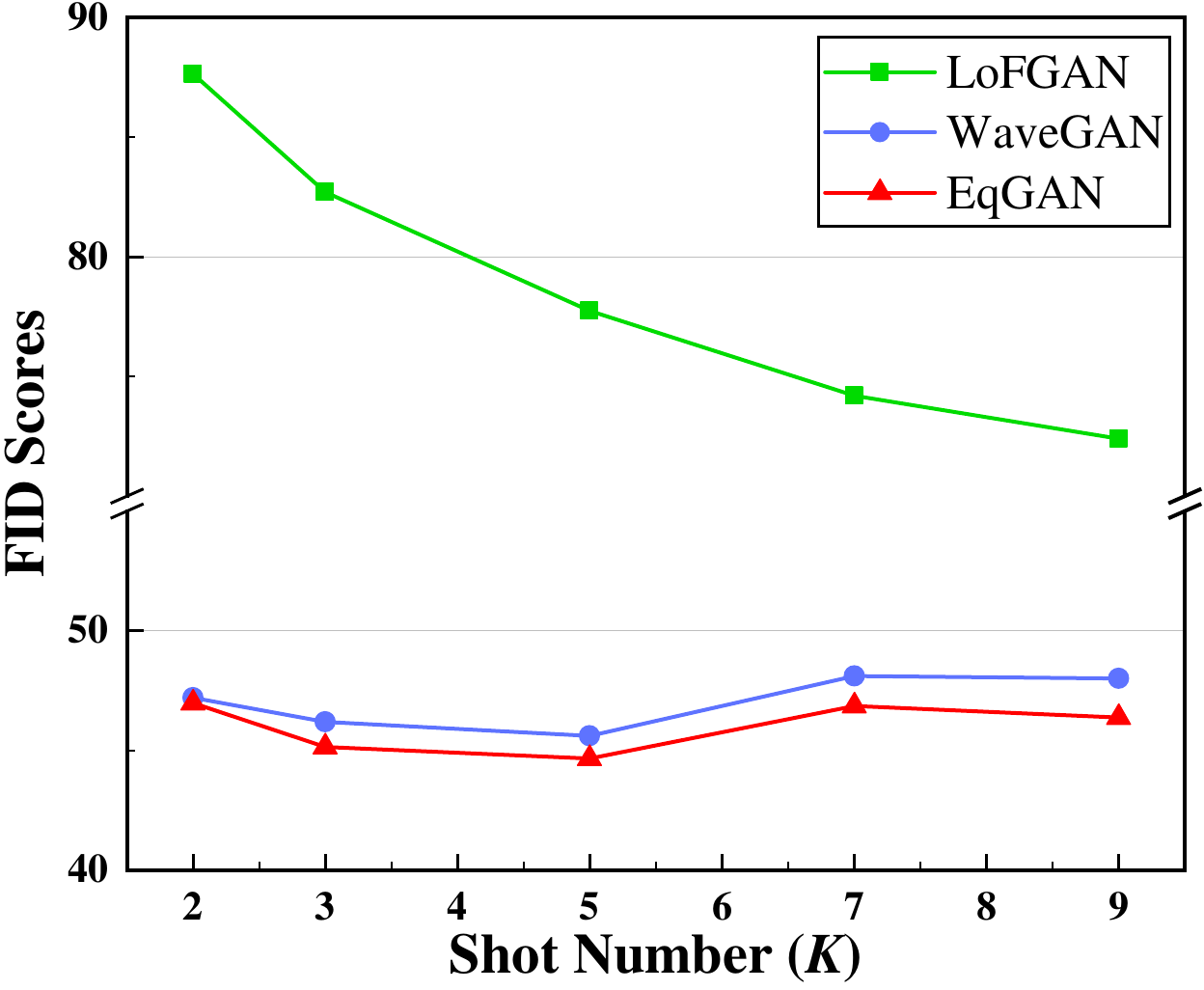}
\end{center}
\vspace{-0.1cm}
   \caption{\textbf{Comparison of FID scores under different shots for few-shot image generation tasks.} Compared to LoFGAN and WaveGAN, EqGAN achieves the lowest FID scores when $K$ increases from $2$ to $9$ for Flower dataset.}
\label{fig:shot}
\vspace{-0.2cm}
\end{figure}
\textbf{Impact of the Number of Shots.}
For the above experiments, all the results are obtained under a 3-shot image generation setting. 
It is worth noting that the number of shots is a critical parameter for few-shot image generation tasks. To investigate the impact of different shots on our model, we compared EqGAN with LoFGAN and WaveGAN under a different number of input images (i.e., $K \in \{ 2, 3, 5, 7, 9 \}$) using the same settings. 
Note that, the number of training and testing images is set to the same for different shots. 
Figure~\ref{fig:shot} shows the FID scores on using different shots for few-shot image generation tasks on Flower dataset. From the figure, we can observe that when $K$ increases from $2$ to $9$, EqGAN achieves the lowest FID scores compared to both
LoFGAN and WaveGAN. The results fully demonstrate that our EqGAN has better adaptability for $K$ and stronger generalization ability for different shot generation tasks. 

\begin{table*}[htbp]
\begin{center}
\setlength{\tabcolsep}{4.5mm}{
\begin{tabular}{lccccccc}
\toprule[1pt]
\multirow{2}{*}{Method} & \multirow{2}{*}{Type} & \multicolumn{2}{c}{Flowers} & \multicolumn{2}{c}{Animal Faces} & \multicolumn{2}{c}{VGGFace} \\
                        &                       & FID $\downarrow$        & LPIPS $\uparrow$        & FID $\downarrow$           & LPIPS $\uparrow$          & FID $\downarrow$        & LPIPS $\uparrow$        \\ \hline
LoFGAN~\cite{Gu0H0021}                & Fusion        & 82.70            & 0.3824              & 113.73               & 0.5020                & 20.61             & 0.3071              \\
EqGAN w/o TS                  & Fusion                & 76.77            & 0.3924              & 76.38               & \underline{0.5093}                & 14.87            & 0.3274              \\
EqGAN w/o SS                 & Fusion                & \textbf{43.75}            & 0.3885              & \underline{37.46}               & 0.5056                & \underline{3.6}            & \underline{0.3315}              \\
EqGAN w/o CE                  & Fusion                & 46.23            & \underline{0.3925}              & 39.33               & 0.5052                & 3.90          & 0.3299              \\
\textbf{EqGAN (ours)}                   & Fusion                &  \underline{45.15}           & \textbf{0.4027}              &  \textbf{34.80}              &  \textbf{0.5130}               & \textbf{3.45}            & \textbf{0.3345}              \\ 
\bottomrule[1pt]
\end{tabular}}
\end{center}
\caption{\textbf{Comparison of quantitative results obtained from ablation studies.} 
The abbreviations of ``TS'', ``SS'' and ``CE'' stand for the texture skip connections, the structure skip connections, and the consistent equalization loss, respectively. The best and the second-best results are \textbf{highlighted} and \underline{underlined}. The symbol $\downarrow$ indicates that lower is better while the symbol $\uparrow$ indicates that higher is better.}
\label{table:abl}
\vspace{-0.2cm}
\end{table*}

\subsection{Ablation Studies (RQ2)}
To evaluate the effectiveness of our fusion strategy, we conducted ablation studies on the feature equalization fusion module and the consistent equalization loss.
For fair comparisons, we used LoFGAN~\cite{Gu0H0021} as a baseline and investigated three main components of our fusion strategy: i) the texture skip connections; ii) the structure skip connections; and iii) the consistent equalization loss. To verify the contribution of each component for EqGAN, we disabled each component and held the other conditions constant. 

\begin{figure}[htbp]
\begin{center}
   \includegraphics[width=1.0\columnwidth]{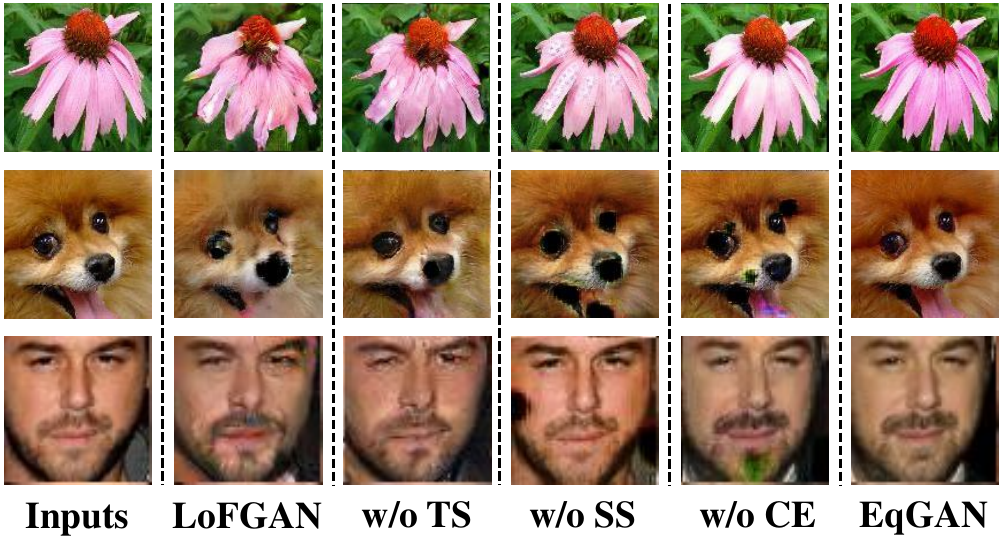}
\end{center}
   \caption{\textbf{Comparison of visualization results for ablation studies.} The left-most column is the input images. LoFGAN is selected as the baseline model. The abbreviations of ``TS'', ``SS'' and ``CE'' denote the texture skip connections, the structure skip connections, and consistent equalization loss, respectively.}
\label{fig:ablation}
\vspace{-0.1cm}
\end{figure}

Table~\ref{table:abl} shows the quantitative results of ablation studies. Compared to the baseline model, each component of EqGAN achieves better FID and LPIPS scores. When we integrated all components, EqGAN achieves the lowest FID scores and highest LPIPS scores on almost all the datasets. 
To further understand
the effectiveness of each component of our method, we presented visualization results of ablation studies in Figure~\ref{fig:ablation}. For the same inputs, the images generated by each ablation scheme are better than the ones generated by LoFGAN on fine-grained texture and structure features with fewer artifacts. Furthermore, we analyzed the role that each component plays in our model. Taking the generated flower images as examples, the flower image generated by ``EqGAN w/o TS'' appears blurry in detail textures while the one generated by ``EqGAN w/o SS'' has artifacts on petals. For the flower image generated by ``EqGAN w/o CE'', there are inconsistent white regions on the generated petals. 
From the above qualitative and quantitative results, we can 
observe the notable 
effectiveness of each proposed component in our 
fusion strategy. Meanwhile,  based on the synergy among TS,
SS and CE, we can find  that EqGAN can achieve the best performance.


\subsection{Augmentation for Classification (RQ3)}
To investigate the performance of generated images for downstream image classification tasks, we performed experiments using
 generated images as augmentations.  For each dataset, 
we divided the unseen categories into three disjoint sets, i.e., 
$\mathbb{D}_{train}$, $\mathbb{D}_{val}$, and $\mathbb{D}_{test}$. 
For the Flower dataset, the data split for $\mathbb{D}_{train}$, $\mathbb{D}_{val}$, and $\mathbb{D}_{test}$ is $10:15:15$ for each category.
For the datasets  Animal Faces and VGGFace, the ratio of each category in $\mathbb{D}_{train}$, $\mathbb{D}_{val}$, and $\mathbb{D}_{test}$ is $30:35:35$. 
For each unseen category, we added $30$ images for the Flower dataset and $50$ for the Animal Faces and VGGFace datasets, respectively. 
Then, we initialized a ResNet18 backbone with seen categories following the same
settings provided by \cite{Gu0H0021} and trained a new classifier using $\mathbb{D}_{train}$.
Here, we used the notation ``Base'' to denote the trained classifier using $\mathbb{D}_{train}$ without any augmentation. 
Meanwhile, we used different few-shot image generation methods to augment $\mathbb{D}_{train}$ to train the classifier and evaluated their resulting classification accuracy using  $\mathbb{D}_{test}$. 

\begin{table}[htbp]
\begin{center}
\begin{tabular}{lccc}
\toprule[1pt]
Method      & Flower  & Animal Faces  & VGGFace \\ \hline
Base         & 61.96       & 20.86             & 49.52        \\
LoFGAN       & 81.18       & 35.71             & 63.80        \\
WaveGAN      & \underline{86.67}       & \underline{56.48}             & \underline{78.35}        \\
\textbf{EqGAN (ours)} & \textbf{88.24}       & \textbf{57.90}             & \textbf{80.32}        \\ \bottomrule[1pt]
\end{tabular}
\end{center}
\caption{\textbf{Top-1 accuracy ($\%$) comparison between  low-data classification  tasks
based on augmentations.} The best
and the second-best results are \textbf{highlighted} and \underline{underlined}, respectively. EqGAN achieves the best accuracy among all four methods.}
\label{table:shot}
\vspace{-0.3cm}
\end{table}

Table~\ref{table:shot} presents the classification results with the augmentation. 
Compared to the results with Base, all the few-shot image generation methods can significantly improve the top-1 classification accuracy. Taking the accuracy on Flower dataset as an example, the performance of LoFGAN, WaveGAN and EqGAN can be improved by 19.22\%, 24.17\% and \textbf{26.28\%}, respectively. 
Note that, EqGAN can achieve the highest accuracy in Flower, Animal Faces, and VGGFace datasets (i.e., \textbf{88.24\%}, \textbf{57.90\%}, and \textbf{80.32\%}), respectively, which outperforms the ones obtained by WaveGAN by \textbf{1.57\%}, \textbf{1.42\%}, and \textbf{1.97\%}, respectively. 
Such improvements in classification accuracy corroborate the benefits of  EqGAN for downstream visual applications. 

\section{Conclusion}

In this paper, we presented a novel fusion-based few-shot image generation framework named \textbf{EqGAN} (feature \textbf{Eq}ualization fusion \textbf{G}enerative \textbf{A}dversarial \textbf{N}etwork), which  explores a more fine-grained fusion strategy with disentangled structures and textures to generate more authentic and diverse images.
Based on our proposed feature equalization module and consistent equalization loss, EqGAN can fuse richer information by using disentangled structures and textures, while strengthening the interaction between the encoder and the decoder to align semantics. 
Comprehensive experimental results on three well-known datasets show that, compared with the state-of-the-art methods, EqGAN can significantly improve not only the quality and diversity of generated images  but also the accuracy of downstream classification tasks. 

\clearpage
{\small

}

\end{document}